\documentclass{article}
\usepackage{spconf,graphicx,multirow}
\usepackage{hyperref}
\usepackage{amsmath}
\usepackage{resizegather}
\usepackage{mathtools}

\title{Supervised Virtual-to-Real Domain Adaptation for Object Detection Task using YOLO}
\name{Akbar Satya Nugraha, Yudistira Novanto, Bayu Rahayudi}
\address{personal.akbarsn@gmail.com, yudistira@ub.ac.id, ubay1@ub.ac.id \\ Informatics Department, Faculty of Computer Science, Brawijaya University, Malang}

\begin{document}

\maketitle

\begin{abstract}
Deep neural network shows excellent use in a lot of real-world tasks. One of the deep learning tasks is object detection. Well-annotated datasets will affect deep neural network accuracy. More data learned by deep neural networks will make the model more accurate. However, a well-annotated dataset is hard to find, especially in a specific domain. To overcome this, computer-generated data or virtual datasets are used. Researchers could generate many images with specific use cases also with its annotation. Research studies showed that virtual datasets could be used for object detection tasks. Nevertheless, with the usage of the virtual dataset, the model must adapt to real datasets, or the model must have domain adaptability features. We explored the domain adaptation inside the object detection model using a virtual dataset to overcome a few well-annotated datasets. We use VW-PPE dataset, using 5000 and 10000 virtual data and 220 real data. For model architecture, we used YOLOv4 using CSPDarknet53 as the backbone and PAN as the neck. The domain adaptation technique with fine-tuning only on backbone weight achieved a mean average precision of 74.457 \%.
\end{abstract}
\begin{keywords}
YOLOv4, Object Detection, Virtual Dataset, Domain Adaptation, Personal Protective Equipment
\end{keywords}
\section{Introduction}
\label{sec:intro}

In the new spring of artificial intelligence, and more specifically in its subfield known as machine learning, a significant number of notable results have shown that usage of machine learning is viable for specific human-task, like object detection and object classification \cite{YOLOv3}. However, a remarkable result of machine learning is also affected by the availability of huge amounts of actual data and its label. 

In the era of big data, having an availability of real input data to train machine learning algorithms is relatively easy for a wide range of applications. Several other fields, however, need more training data. Even though data is available, it must be manually revised to make it usable.

Making a dataset usable for training is complicated and needs technical knowledge, especially datasets, for object detection. This is because object detection is needed an object anchor as the label for every image. Training an anchor-based object detector with a sparsely annotated dataset can cause performance degradation \cite{Yoon2021}. 

A problem like the availability of datasets and the taxing process for revising data to make it usable force researchers to find another method. Among the methods of leveraging trained model, synthetic data, computer-generated datasets, or virtual datasets are used for pre-training dataset. Virtual datasets have been on the rise as they offer an abundant data scenario and correctly label it at a lower cost.   

The downside of using a virtual dataset comes with a problem: cross-domain shift. Cross-domain object detection is challenging due to multi-level domain shift in an unseen domain \cite{Liu2021HDCN}. Research has already been conducted and shown a few methods for solving cross-domain shifts. It varies from adding a domain-adapting layer \cite{Liang2020} or creating a hierarchical domain-consistent network \cite{Liu2021HDCN} to solving cross-domain shift problems for using virtual data. 

This research investigates a domain adaptation strategy that maximizes the utilization of the virtual domain in the real-world domain. Hence, an object detection model needs fewer data for the real-world domain. Specifically, we demonstrate how the transfer learning approach on a well-known deep neural network can achieve state-of-the-art results in automatic visual media indexing after being trained with virtually generated images of people wearing safety gear, such as high-visibility jackets and helmets and domain adaptation using a few real image training examples.

\section{Related Work}
\label{sec:format}

Object detection technologies achieved amazing accuracies with faster, unimaginable speeds a few years ago. Recently, YOLO \cite{Redmon2016} \cite{Bochkovskiy2020} \cite{Redmon2018} and RCNN \cite{Ren2015} are de facto standard for object detection tasks. Most of the research on object detection is huge generic annotated datasets, such as Pascal \cite{Pascal}, ImageNet \cite{ImageNet}, MS COCO\cite{MSCoco}, or OpenImages\cite{OpenImages}. This dataset collects a large number of manually annotated web images.

With the need for huge amounts of data to reach reliable accuracy, virtually computer-generated or virtual datasets gained significant interest. Usage of virtual dataset begins from research to detecting pedestrians using the virtual dataset, which shows promising results with less than 2\% derivation rate for detecting pedestrians \cite{Marin2010}. The virtual dataset was also used to study trained CNNs to qualitatively and quantitatively analyze deep features \cite{Aubry}.

The usage of data generated from the game was also explored in a few research. In \cite{martinez2017}, using 50000 labeled images from the GTA-V game trained on CNN shows that the mean squared error for lane distance estimation is considerably small for only a virtual dataset. In \cite{Qiu2016} shown using the unreal engine, an RCNN model could detect a sofa from a different viewpoint by only using a dataset generated from Unreal Engine 4

Dataset from GTA-V demonstrated that it is possible to reach excellent results on tasks such as real people tracking and pose estimation \cite{Fabbri2018}. Using Faster R-CNN on virtual datasets and validating the result on the KITTI dataset also shows good results \cite{JohnsonRoberson2016}. The virtual dataset could also be used to train a simple convolutional network to detect objects belonging to various classes in video \cite{Bochinski2016}.

Object detection models could also use the virtual dataset to achieve better accuracy. For example, in \cite{martinez2017} using virtual dataset as SIM 10k into real dataset Cityscapes for car detection, resulting average precision 51.6\%. Using 140.000 virtual images and just 220 images resulting an object detection model that could detect Personal Protective Equipment (PPE) with 76\% accuracy \cite{diBenedetto2019}. Based on the research above, using a virtual dataset could create a better-accuracy model.

\section{Methodology}
\label{sec:pagestyle}

\subsection{Virtual Data}
\begin{figure}[htb]

\begin{minipage}[b]{.48\linewidth}
  \centering
  \centerline{\includegraphics[width=4.0cm]{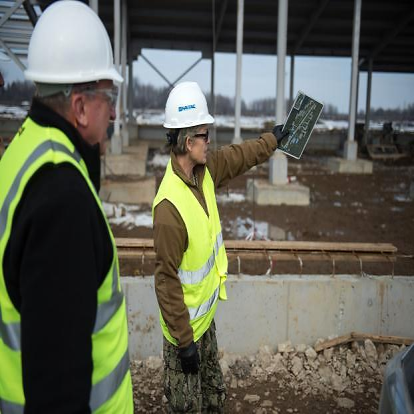}}
  \centerline{(a) Image for real dataset}\medskip
\end{minipage}
\hfill
\begin{minipage}[b]{0.48\linewidth}
  \centering
  \centerline{\includegraphics[width=4.0cm]{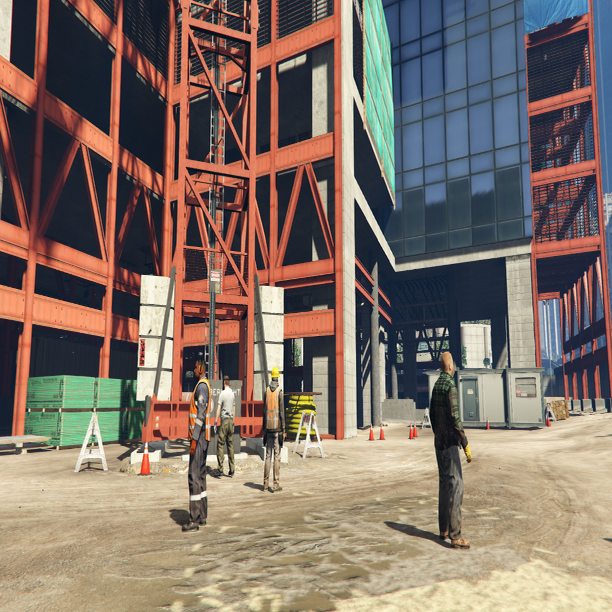}}
  \centerline{(b) Image for virtual dataset}\medskip
\end{minipage}
\caption{Image sample for real \& virtual dataset}
\label{fig:res}
\end{figure}

We used the VW-PPE dataset with over 140.000 virtual and 220 real images. The virtual images were generated using RAGE, the game engine for GTA-V, with each image having a width of 1088 and a height of 612, but for real images, each image has different width and height. VW-PPE dataset has seven object classes: Bare Head, Helmet, Ear Protection, Welding Mask, Bare Chest, High Visibility Vest, and Person. The virtual images have been generated in 10 locations of the game map, with three weather and time variations for each location. From 140000 virtual images, this research only used 5000 and 10000 using random sampling. Real images will be split by 50:50 for training and testing dataset. Sample images in the VW-PPE dataset are shown in Fig. 1.

\subsection{YOLO}
\begin{figure}[htb]

\begin{minipage}[b]{1.0\linewidth}
\centering
\centerline{\includegraphics[width=8.5cm]{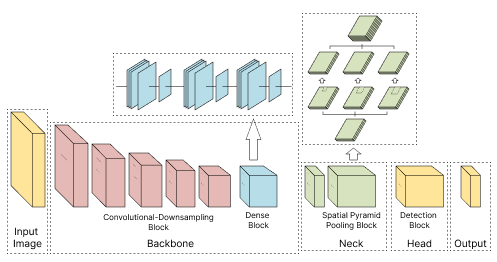}}
\end{minipage}

\caption{YOLOv4 architecture}
\label{fig:res}
\end{figure}
We used the architecture for object detection,  You Only Look Once (YOLO), a one-stage detector that could do image localization and classification in one stage. We used it for the object detection task YOLOv4. YOLOv4 is used for this research because of the custom component that could be used in YOLOv4. Architecture YOLOv4 using CSPDarknet53 \cite{CSPDarknet53} as the backbone, PAN \cite{PAN} as neck, and YOLOv3 detector layer \cite{YOLOv3} as the head.

To evaluate the performance of our implementation, we used Intersection over Union (IoU) based on the area of the detected (D) and real (V) bounding boxes, as well as Precision (Pr) and Recall (Rc). The confidence score associated with detected bounding boxes varies from 0 to 1. They are included in the output if their confidence score exceeds a user-defined threshold. Given the preceding criteria, the mean Average Precision (mAP) is calculated as the average of the highest precision at various recall settings.

\subsection{Loss Function}

To achieve robust detection from training a machine learning model, we used YOLOv4 loss function. The first component of YOLOv4 loss function is the Complete Intersection over Union (CIoU) loss formula to compute loss using x and y coordinates of the width and height of the bounding boxes \cite{Zheng2019}.

\begin{equation}
\label{eq:alpha}
\alpha = \frac{\upsilon}{(1 - IoU) + \upsilon^{'}}
\end{equation}

\begin{equation}
\label{eq:upsilon}
\upsilon = \frac{4}{\pi^{2}} (arctan \frac{w^{gt}}{h^{gt}} - arctan\frac{w}{h})^{2}
\end{equation}

Inside CIoU formula, there are two variables, that is \begin{math} \alpha \end{math} of a positive trade-off parameter, explained in Equation \ref{eq:alpha} and \begin{math} \upsilon \end{math} of the consistency of aspect ratio, explained in Equation \ref{eq:upsilon}. The formula of \begin{math} L_{CIoU} \end{math} is explained in Equation \ref{eq:CIoU}.

\begin{equation}
\label{eq:CIoU}
L_{CIoU} = \left [ 1 - IoU + \frac{\rho(b,b^{gt})}{c^{2}} + \alpha \upsilon \right ]
\end{equation}



\begin{equation}
\label{eq:yolo}
\begin{multlined}
L_{total} = L_{CIoU} \\
- \sum_{i=0}^{S^{2}}\sum_{j=0}^{B} I_{ij}^{obj}\left [ \hat{C_{i}} log(C_{i}) + (1 - \hat{C_i}log(1-C_{i})) \right ] \\
- \lambda_{noobj}\sum_{i=0}^{S^{2}}\sum_{j=0}^{B} I_{ij}^{noobj}\left [ \hat{C_{i}} log(C_{i}) + (1 - \hat{C_i}log(1-C_{i})) \right ] \\
- \sum_{i=0}^{S^{2}} I_{ij}^{obj}\sum_{c \in classes} \left [ \hat{p}_{i}(c) log (p_{i}(c)) + (1 - \hat{p}_{i}(c)log(1 - p_{i}(c))) \right ]\end{multlined}
\end{equation}

Inside Equation \ref{eq:yolo}, second and third components were calculated as the confidence scores of objectness inside every grid cell. The variable of \begin{math} I_{ij}^{noobj} \end{math} and \begin{math} I_{ij}^{obj} \end{math} show the presence and absence of an object on that pixel, respectively. Value of \begin{math} I_{ij}^{obj} \end{math} will be 1 if there are objects in the grid cell, and \begin{math} I_{ij}^{noobj} \end{math} will be 1 if there is no object in the grid cell and 0 conversely. The variable of \begin{math} C_{i}\end{math} and \begin{math}\hat{C}_{i} \end{math} are confidence scores of ground truth and prediction of whether there is an object or not, respectively. At the last component, there are \begin{math} \hat{p}_{i}\end{math} and \begin{math} p_{i} \end{math} variables of actual and prediction class, respectively, for classification loss.

\subsection{Domain adaptation}
\begin{figure}[htb]

\begin{minipage}[b]{1.0\linewidth}
\centering
\centerline{\includegraphics[width=8.5cm]{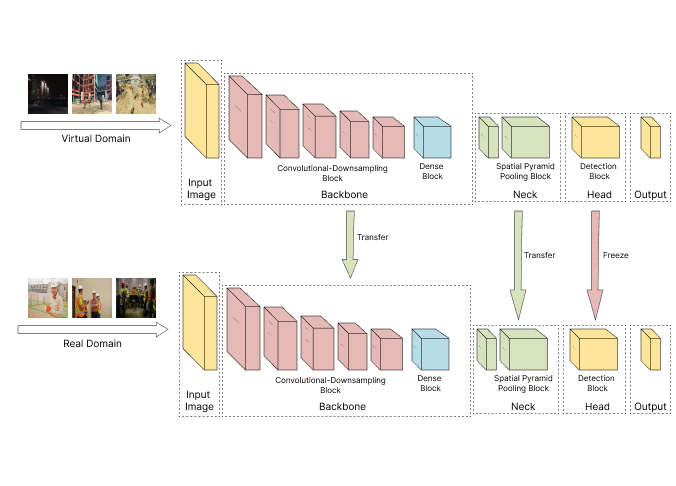}}
\end{minipage}

\caption{SHOT Domain Adaptation Scheme}
\label{fig:res}
\end{figure}
We proposed using domain adaptation by pre-training virtual datasets to solve cross-domain shift problems during tuning. Specifically, we apply the domain adaptation method to adapt pre-trained YOLO to our case. Our premise is that a pre-trained network contains sufficient knowledge for us to specialize it for a task using the transfer learning capabilities of deep neural networks and training sets generated from the virtual world.

The objective of transfer learning is to utilize the first already trained layers (i.e., those identifying low-level features) and update the final layers of the network to expand the detection capabilities to the new set of objects. With a trained deep convolutional neural network, its first layers have learned to identify increasingly complex features.

We used a domain adaptation scheme based on SHOT (Source Hypothesis Transfer) \cite{Liang2020} in this experiment. We explained this in Fig. 3. For addressing the domain shift problem, we implemented the SHOT Domain Adaptation Scheme, where the last layer of the YOLO architecture utilized for detecting bounding boxes would be frozen. In addition to the weight of the freezing detecting layer, we will transfer the weight of the backbone and neck.


\section{Experiments}
\label{sec:typestyle}

This scheme is explained in Fig. 2. We trained 6 schemes, that is as below:
\begin{itemize}
\item Training from scratch using real dataset only (YR)
\item Transfer learning from scratch (YVR)
\item Transfer learning with pre-trained weight (YCVR)
\item Transfer learning with domain adaptation scheme (YCSVR)
\item Transfer learning with mosaic augmentation and pre-trained weight (YCMVR)
\item Transfer learning with only backbone weight and mosaic augmentation (YCMSVR)
\end{itemize}
\begin{table}[]
\begin{center}
\begin{tabular}{|c|l|l|}
\hline
\textbf{Scheme} & \multicolumn{1}{c|}{\textbf{Total Sample Data}} & \multicolumn{1}{c|}{\textbf{mAP}} \\ \hline
YR & 220 & 0 \\ \hline
\multirow{2}{*}{YVR} & 5000 & 27.251 \\ \cline{2-3}
& 10000 & 51.369 \\ \hline
\multirow{3}{*}{YCVR} & 5000 & 65.513 \\ \cline{2-3}
& 10000 & 72.264 \\ \cline{2-3}
& 20000 & 59.691 \\ \hline
\multirow{3}{*}{YCSVR} & \textbf{5000} & \textbf{74.457} \\ \cline{2-3}
& 10000 & 72.096 \\ \cline{2-3}
& 20000 & 73.369 \\ \hline
\multirow{2}{*}{YCMVR} & 5000 & 55.010 \\ \cline{2-3}
& 10000 & 54.368 \\ \hline
\multirow{2}{*}{YCMSVR} & 5000 & 59.977 \\ \cline{2-3}
& 10000 & 53.788 \\ \hline
\end{tabular}
\caption{mAP result from all testing scheme}
\end{center}
\end{table}

Based on Table 1, YR receives 0 mAP, since no detections achieved the confidence level. Utilizing 5000 sample data, the mAP for YVR hits 27.251. By using 10000 sample data, the mAP reaches 51.369. Using virtual datasets as source domains before transferring learning to real-world datasets is a promising strategy for boosting mAP in object detection tasks, as demonstrated by these results.

YCVR outperforms YVR, where the mAP for 5000 sample data is 65.515, and for 10000 sample data, the mAP is 72.264. Fine-tuning pre-trained weight, even if it is cross-domain, increases the mAP for the object identification model based on this finding.

With 5000 virtual sample data, YCSVR achieves the best mAP score of 74.457; while utilizing 10,000 virtual sample data, the mAP score hits 72.096. Based on these findings, it appears that transfer learning utilizing the SHOT Domain Adaptation Scheme will increase mAP, however, it will struggle when the proportion of virtual domain data is considerably higher than real domain data.

Lastly, with YCMVR and YCMSVR, it is demonstrated that mosaic augmentation decreases mAP. All YCMVR and YCMSVR tests reveal a mAP between 50 and 59, which is lower than YCVR.

\begin{table}[]
\begin{center}
\begin{tabular}{|c|l|}
\hline
\textbf{Class} & \multicolumn{1}{c|}{\textbf{AP}} \\ \hline
Head & 84.052 \\ \hline
\textbf{Helmet} & \textbf{93.691} \\ \hline
\textbf{Ear Protection} & \textbf{42.292} \\ \hline
Welding Mask & 86.364 \\ \hline
Bare Chest & 59.159 \\ \hline
High Visibility Vest & 87.637 \\ \hline
Person & 51.457 \\ \hline
\end{tabular}
\caption{Average precision of each class using best scheme}
\end{center}
\end{table}

\begin{figure}[htb]

\begin{minipage}[b]{1.0\linewidth}
\centering
\centerline{\includegraphics[width=2.8cm]{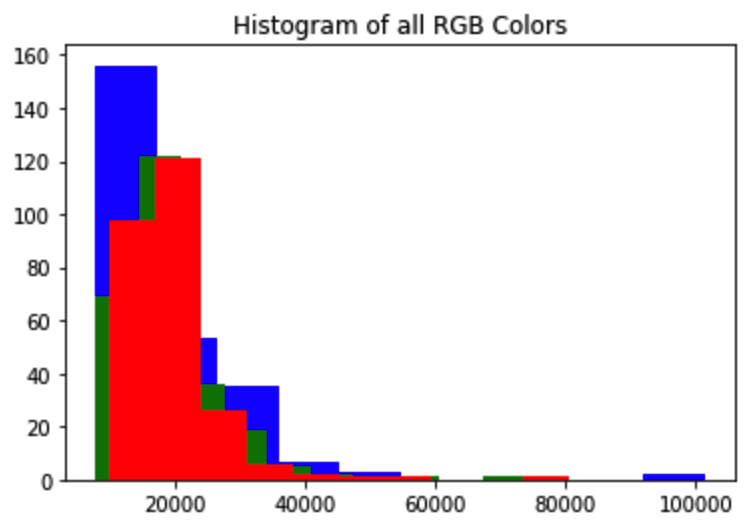}}
\centerline{(a) Real Dataset}\medskip
\end{minipage}
\hfill
\begin{minipage}[b]{.48\linewidth}
\centering
\centerline{\includegraphics[width=2.8cm]{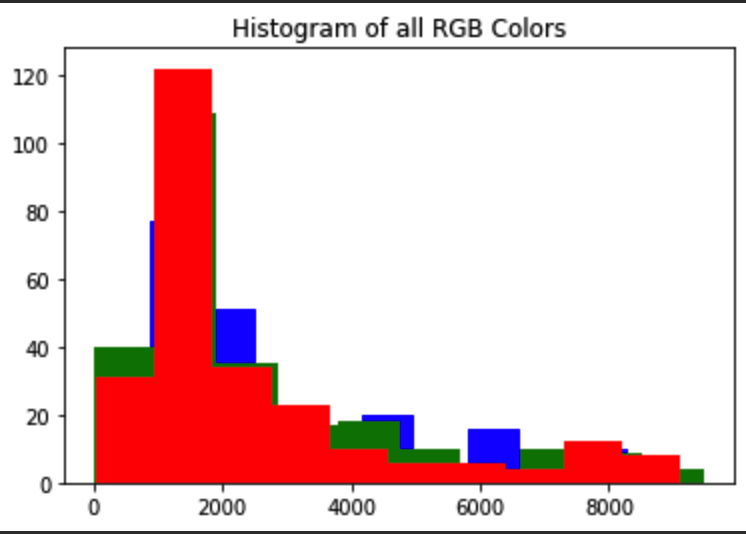}}
\centerline{(b) Sample 5.000}\medskip
\end{minipage}
\hfill
\begin{minipage}[b]{0.48\linewidth}
\centering
\centerline{\includegraphics[width=2.8cm]{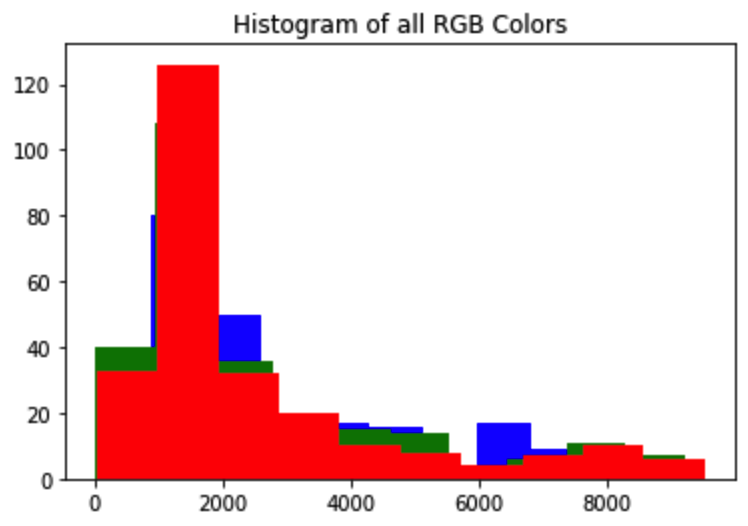}}
\centerline{(c) Sample 10.000}\medskip
\end{minipage}
\caption{Average color histogram from all dataset used in research}
\label{fig:res}
\end{figure}

Table 1 shows that mAP from YCSVR using 10.000 training data is lower than sample 5.000. This is because the sampling process is random. Although class distribution is in the same ratio, the image is still different. Fig. 4 shows that the real dataset is brighter than the 2 sample data in the virtual dataset. The issue with randomly sampling virtual datasets is that the average histogram color of each sampled virtual dataset will be darker than that of the actual dataset. Therefore, the domain shift problem can be caused by random sampling, which makes virtual datasets darker than real datasets.

Table 2 explains the average precision of every class using the best scheme, YCSVR. It shows that the helmet class has the highest average precision, and the ear protection class has the lowest average precision. This is because the helmet class has the most class label in the dataset, while ear protection has the fewest class label.

\begin{figure}[htb]

\begin{minipage}[b]{.48\linewidth}
\centering
\centerline{\includegraphics[width=4.0cm]{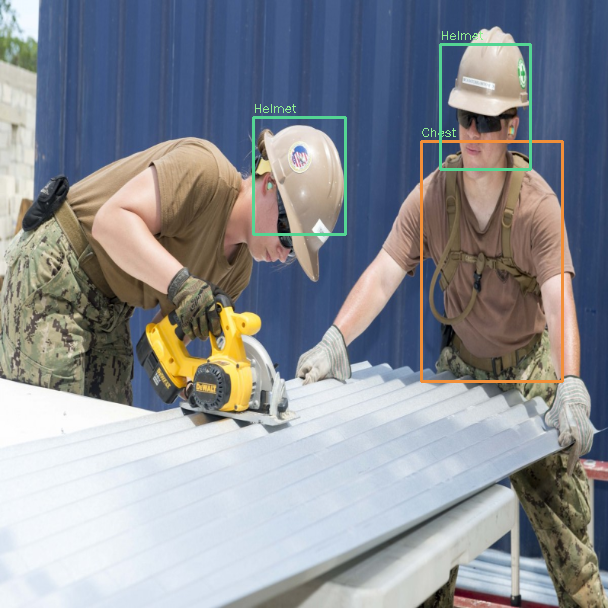}}
\end{minipage}
\hfill
\begin{minipage}[b]{0.48\linewidth}
\centering
\centerline{\includegraphics[width=4.0cm]{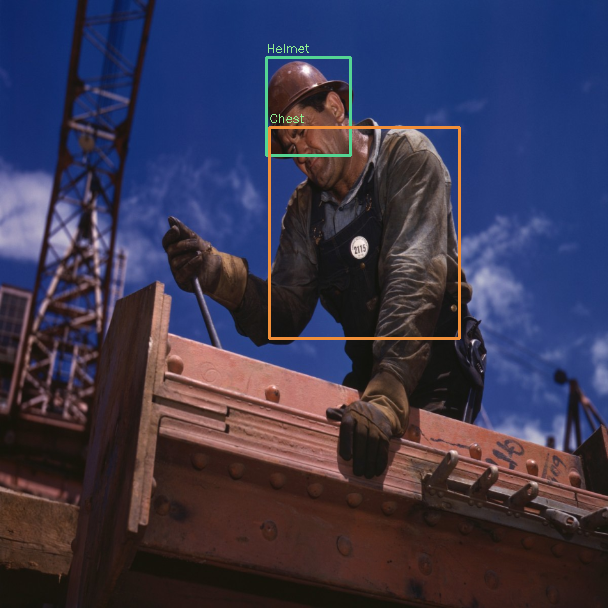}}
\end{minipage}
\caption{Sample of detection of PPE objects}
\label{fig:res}
\end{figure}

Using YCSVR models has a promising result for both bounding box prediction and class classification, as shown in Fig. 5.

\section{Conclusion}
\label{sec:majhead}
Training a deep neural network in virtual environments has been proven to help when the number of available and usable training datasets is low. In this paper, we performed personal protective equipment object detection with a few real data/images.
In our experiment, we trained YOLOv4 on the virtual dataset and tested it on a real dataset. In addition, we also fine-tune the deep neural network with small real data. Based on the experiment, we found that the performance of transfer learning only backbone weight is better than normal transfer learning. Moreover, we found that there are better choices than mosaic augmentation for training object detection cross-domain.

\vfill\pagebreak

\bibliographystyle{IEEEbib}
\bibliography{export}

\end{document}